\documentclass{article}
\usepackage{arxiv}
\usepackage[utf8]{inputenc} 
\usepackage{hyperref}       
\usepackage{nicefrac}       
\usepackage{microtype}      
\usepackage{lipsum}
\usepackage{times}
\usepackage{latexsym}
\usepackage{url}
\usepackage[T3,OT2,T1]{fontenc} 
\usepackage{caption}
\usepackage[noenc]{tipa}
\usepackage{tikz}
\usepackage{enumitem}
\usepackage{graphicx}%
\usepackage{multirow}%
\usepackage{amsmath,amssymb,amsfonts}%
\usepackage{amsthm}%
\usepackage{mathrsfs}%

\usepackage[figuresright]{rotating}%
\usepackage{xcolor}%
\usepackage{textcomp}%
\usepackage{manyfoot}%
\usepackage{booktabs}%
\usepackage{algorithm}%
\usepackage{algorithmicx}%
\usepackage[noend]{algpseudocode}%
\usepackage{program}%
\usepackage{listings}%
\usepackage{subcaption}
\algnewcommand{\IIf}[1]{\State\algorithmicif\ #1\ \algorithmicthen}
\algnewcommand{\EndIIf}{}
\algnewcommand{\FFor}[1]{\State\algorithmicfor\ #1\ \algorithmicdo}
\algnewcommand{\EndFFor}{}

\title{An Attention Matrix for Every Decision: Faithfulness-based Arbitration Among Multiple Attention-Based Interpretations of Transformers in Text Classification}

\author{
  Nikolaos Mylonas\\
  Aristotle University of \\Thessaloniki, 54636, Greece\\
    \texttt{myloniko@csd.auth.gr}\\
     \And
       Ioannis Mollas\\
  Aristotle University of \\Thessaloniki, 54636, Greece\\
             \texttt{iamollas@csd.auth.gr}\\
             \And
               Grigorios Tsoumakas\\
  Aristotle University of \\Thessaloniki, 54636, Greece\\
             \texttt{greg@csd.auth.gr}\\
}

\begin{document}

\maketitle

\begin{abstract}
Transformers are widely used in natural language processing, where they consistently achieve state-of-the-art performance. This is mainly due to their attention-based architecture, which allows them to model rich linguistic relations between (sub)words. However, transformers are difficult to interpret. Being able to provide reasoning for its decisions is an important property for a model in domains where human lives are affected. With transformers finding wide use in such fields, the need for interpretability techniques tailored to them arises. We propose a new technique that selects the most faithful attention-based interpretation among the several ones that can be obtained by combining different head, layer and matrix operations. In addition, two variations are introduced towards (i) reducing the computational complexity, thus being faster and friendlier to the environment, and (ii) enhancing the performance in multi-label data. We further propose a new faithfulness metric that is more suitable for transformer models and exhibits high correlation with the area under the precision-recall curve based on ground truth rationales. We validate the utility of our contributions with a series of quantitative and qualitative experiments on seven datasets.
\end{abstract}

\keywords{Interpretable Machine Learning \and Transformers \and Attention \and Text Classification \and Multi-Label Learning}

\section{Introduction}

Transformers have become the dominant approach for tackling natural language processing (NLP) tasks, surpassing previous convolutional and recurrent neural network architectures~\cite{vaswani2017attention,wolf-etal-2020-transformers}. 
On the other hand, transformers are black boxes. The large number of their parameters and their complex architecture makes it difficult to understand how they reach their decisions~\cite{transformer_not_inter}. Interpretability is important in high-risk applications, where decision-making systems can have a significant impact on human lives~\cite{highrisk2}. The importance of interpretability in such applications, along with the high performance that transformers can achieve in them, raises the need for techniques that can explain the decisions of these models.

The most popular transformer-specific interpretability approach is the use of self-attention scores. Since these scores are computed during inference, obtaining interpretations from them adds no computational overhead. However, the use of attention to produce explanations has been met with skepticism by some researchers~\cite{notAtt}. Other transformer-specific interpretability approaches combine attention with gradient information~\cite{chefer2021transformer} or compute new attentions based on the network's residual connections~\cite{DBLP:journals/corr/abs-2005-00928}. Nevertheless, such techniques introduce new elements in the model's architecture, necessitating the model's pre-training from scratch.


While reviewing the literature on attention-based interpretation of transformers, we ran across different ways of integrating attention information across heads and layers, as well as different ways of extracting an interpretation vector from a final attention matrix. This motivated us to study whether some particular ways perform better than the rest. The inconclusiveness of this study led us to propose a novel family of local interpretation techniques for transformers, dubbed \textsc{Optimus}. Given an input instance, \textsc{Optimus Prime} evaluates the faithfulness of the interpretations obtained by a number of different combinations of head, layer and matrix operations and selects the best one as the final interpretation. In multi-label learning tasks, \textsc{Optimus Label} selects the most faithful combination separately for each predicted label, leading to improved results. To reduce the computational complexity, at the cost of performance, \textsc{Optimus Batch} selects the most faithful combination across an initial set of instances, and then uses this fixed setup for subsequent instances. In addition, we propose a new faithfulness evaluation metric, {\em Ranked Faithful Truthfulness}, that correlates highly with the area under the precision recall curve computed on top of ground truth rationales.


Our contributions are empirically assessed on seven datasets from four different domains: sentiment analysis, natural language understanding, hate speech detection, and biomedicine. Decision-making in the latter two can in some cases have significant impact on human lives. The first one by affecting freedom of speech or allowing the incitement of violence, and the second one by recommending inappropriate treatments. Results show that attention-based interpretations can compete with state-of-the-art techniques, and even exceed them in certain cases, while being less computationally expensive in others. 

The remainder of this article is organized as follows. Section~\ref{sec:related-work} examines related work on transformer interpretability and interpretability evaluation. Section~\ref{sec:opt} presents our attention-based interpretability technique alongside our new faithfulness metric. Section~\ref{sec:exp} introduces the setup of our experiments and presents quantitative and qualitative evaluation results. Finally, Section~\ref{sec:con} discusses the conclusions of our work and points to future research directions.

\section{Related Work}
\label{sec:related-work}

A model's ability to provide insights for its decisions or inner working, whether intrinsically or not, is referred to as interpretability. Complex models, such as transformers, cannot provide interpretations out of the box, and therefore post-hoc techniques are typically applied. The representations of an interpretation include, among others, rules, heatmaps, and feature importance. This work focuses on feature importance, also known as attribution importance or saliency map, which quantifies the influence of a model's input features on its output. The rest of this section presents related work on transformer interpretability, mainly in the context of text classification, as well as on interpretability evaluation methods.

\subsection{Transformer interpretability}

We first review model agnostic and neural-specific feature importance techniques that are applicable to transformers. Then, we present interpretability techniques that have been designed specifically for transformers. Finally, we discuss two studies that have discovered interesting patterns and properties of transformers' attention module. 

\subsubsection{Transformer-applicable techniques}
LIME~\cite{LIME} and SHAP~\cite{SHAP}, two model-agnostic, local interpretation approaches, can be easily applied to a transformer by just probing it for predictions. Backpropagation-based neural-specific techniques such as Layer-wise Relevance Propagation (LRP)~\cite{LRP} and Integrated Gradients (IG)~\cite{IG} can be modified to provide interpretations for transformer models. Such techniques that consider model architecture and employ back-propagated gradients are expected to yield more meaningful interpretations. However, some studies have shown that model-agnostic methods achieve competitive performance in transformer explainability~\cite{hateXplain}. 

Making use of techniques such as LIME, IG, and SHAP, Thermostat~\cite{Feldhus2021ThermostatAL} provides a collection of ready-to-use interpretations in the form of feature importance scores for different transformer models and datasets. This can reduce the environmental impact and economic barriers associated with the repetitive execution of common experiments in interpretable NLP. Ecco~\cite{alammar-2021-ecco} is an open-source library offering a variety of techniques for analyzing the inner workings of a transformer, such as how the model's hidden states change from layer to layer, providing feature importance interpretations, as well as enabling the examination of activation vectors.

\subsubsection{Transformer-specific techniques}

Extracting information from the attention module of transformers has been a popular method for interpreting their decisions~\cite{attExp1,attExp2}, especially before its criticism~\cite{notAtt,notAtt2}. Recent work introduced an interpretation technique based on reinforcement learning that uses attention matrices in order to build a perturbation-based game environment that provides explanations for transformer models~\cite{ijcai2022p102}. An explainability method based on hierarchical transformer models was proposed in~\cite{electronics10182195}. Two transformer-based model architectures were introduced to classify and extract explanations for sentiment analysis. Explanations were extracted based on attention weights and compared to ones provided by human users. 

A recent method that does not solely rely on raw attention to provide explanations, is combining relevance and gradient information~\cite{chefer2021transformer}. Specifically, relevance scores are produced for each attention head in each layer, leveraging the theory underpinning LRP. These results are then integrated with gradient information. The produced explanation is a matrix of size $S \times S$, where $S$ denotes sequence length. The final relevance map is derived from the row of the matrix corresponding to the [CLS] token.

A process aimed at quantifying how attention information flows from layer to layer is introduced in~\cite{DBLP:journals/corr/abs-2005-00928}. Specifically, two methods based on Directed Acyclic Graphs are proposed, {\em Attention Rollout} and {\em Attention Flow}, that compute attentions for each input token. Both methods take into account the models' residual connections to obtain token attentions. These attentions were found to retain more information and can serve as a visualization tool. A tool developed specifically for attention visualization, is BertViz~\cite{DBLP:journals/corr/abs-1906-05714}, which provides insight about how tokens of a particular sentence affect each other, while also shedding light on what each attention head and layer focuses on. 

\subsubsection{Attention analysis in transformers}
Five distinct patterns of self-attention that are used across attention heads were discovered in~\cite{DBLP:journals/corr/abs-1908-08593} by displaying attention score heatmaps for BERT. Additionally, it was shown that by disabling certain attention heads or layers, the model does not necessarily display a decrease in performance and can even exhibit improvements in specific cases.

The topic of transformer identifiability was explored in~\cite{DBLP:conf/iclr/BrunnerLPRCW20}. Attention weights are defined as identifiable if they can be uniquely determined from the transformer's output. It was discovered that if the sequence length used is higher than the attention head dimension, then attention weights are not identifiable. This is because certain rows of the attention matrix can be linear combinations of others. Using attention as an explanation may be unwarranted, since different combinations of weights may produce the same output.

\subsection{Interpretability evaluation}
\label{mezures}

The most appropriate way to evaluate an interpretability technique is via a user study, where end users compare interpretations~\cite{evrikaKatharistiko}. However, this kind of experimental procedure is not always feasible, due to its costly and time-consuming nature. Furthermore, human evaluation is prone to bias~\cite{persilMalaktiko}. 

Ground truth interpretations provided by human annotators are called {\em rationales}~\cite{eraser}. In text classification, these rationales can be words, sentences or spans of text that strongly associate each instance with its label. When rationales are available, we can evaluate interpretations using standard metrics, such as $F_1$, or the area under the precision-recall curve (AUPRC). Unfortunately, datasets accompanied by rationales are scarce. Moreover, since rationales are provided by humans, erroneous, noisy, and biased annotations may occur, as in human evaluation.

There are also metrics that can do without human input, by evaluating certain properties of the produced interpretations. {\em Robustness}~\cite{robustness} concerns the stability of a technique. By slightly modifying the examined instances, robustness measures the degree of change between the interpretations for the initial and modified instances. The smaller this change is, the higher the robustness of the technique. {\em Comprehensibility}~\cite{comprehensibility_nzw} calculates the percentage of non-zero weights in an interpretation. The lower this number, the easier for end users to comprehend the interpretation.

A frequently used family of metrics are those emulating the behavior of a user that interacts with the model to explore the validity of a given interpretation, known as faithfulness evaluation metrics. {\em Faithfulness score}~\cite{faithfulness}, the most popular one, eliminates the token with the highest importance score from the examined instance and measures how much the prediction changes. Higher changes signify better interpretations. {\em Truthfulness}~\cite{mollas_lionets_2022} removes all tokens of an instance, one at a time, and awards or penalizes the technique based on the model's behavior. Other metrics of this family include Comprehensiveness, Sufficiency, Monotonicity and Faithfulness Violation Test~\cite{chan-etal-2022-comparative,liuLGKL022}.

\section{Our Approach}
\label{sec:opt}

Given a transformer model $f$, and an input sequence $x = [t_{1}, \dots, t_{S}]$, consisting of $S$ tokens $t_i, i = 1 \ldots S$, our goal is to extract a local interpretation $z= [w_{1}, \dots, w_{S}]$, where $w_{i}\in \mathbb{R}$ signifies the influence of token $t_{i}$ on the model's decision $f(x)$, based on the model's self-attention scores. We first present the \textsc{Optimus} family of techniques for selecting the most faithful interpretation among several different ones. Then, we discuss a novel faithfulness metric that exhibits high correlation with AUPRC computed on top of ground truth rationales. 

\subsection{The \textsc{Optimus} family of techniques}

We first review the self-attention layer of transformers, where attention scores are computed, and introduce a variation for obtaining feature importance scores that include negative values. Then, we present the different ways that are commonly used to turn attention scores into an interpretation in the form of feature importance, as well as introduce some new ones. Finally, we present three techniques for selecting the most faithful among these interpretations.

\subsubsection{Attention scores}

The input to each self-attention layer is a matrix of dimensions $S \times E$, where $S$ denotes sequence length, and $E$ refers to embedding size. At first, this matrix is passed through three linear layers, namely \textit{Query}, \textit{Key}, and \textit{Value}, to produce matrices $Q$, $K$, $V$ of the same dimension as the input. Next, the dot product of $Q$ and $K$ is calculated, and divided by the square root of the embedding size. Subsequently, the attention mask is added, and the result is then passed through a softmax function, which outputs a matrix, $A$, of dimensions $S \times S$, containing the attention from each token of the sequence to the rest: 

\begin{equation}
A = softmax(\frac{Q \cdot K^{T}}{\sqrt{E}}+mask)
    \label{eq:attention_score}
\end{equation}

In fact, transformers employ a \textit{multi-head attention} architecture, where the input to each self-attention layer is logically split to $R$ attention heads. Each head operates on a different part of the input matrix of size $S \times \frac{E}{R}$ allowing the transformer to learn different relationships between tokens. Due to the use of the softmax function, attention matrices contain only positive numbers. Consequently, any interpretations extracted from these matrices will contain only positive values. However, interpretations containing both positive and negative values are often desirable~\cite{9671639}, as the presence of polarity within feature importance scores can facilitate a better association of input elements to a decision. Therefore, our experiments consider a modification on Eq.~\ref{eq:attention_score}, which ignores the softmax function to allow negative values to appear in the resulting interpretations. We denote the corresponding matrix by $A^*$.




\subsubsection{Interpretation extraction}

An interpretation is typically obtained by first aggregating the attention matrices across all heads of each self-attention layer, then aggregating the resulting matrices across all self-attention layers, and finally extracting from the resulting matrix the one-dimensional interpretation vector. 
Head operations commonly found in the literature are averaging~\cite{chefer2021transformer,hateXplain,wanktree} and summing~\cite{exbert,schwenke2021show} the attention matrices of each head. These operations are equivalent in the context of interpretability evaluation, as they lead to the same ordering of the tokens by importance, differing only in the magnitude of the scores assigned to the tokens. Therefore, the summing operation is not included in our pipeline. Operations concerning the resulting matrices of self-attention layers include averaging~\cite{schwenke2021show} and multiplying~\cite{chefer2021transformer}. We therefore also consider multiplying as an operation for heads.   

We further propose an additional operation for both heads and layers: selection of the attention matrix corresponding to a certain head/layer. As different heads are learning different relationships among the input tokens, including the [CLS] token in classification tasks, they are essentially learning different ways that input tokens influence the class. We therefore hypothesize that this new selection operation can be crucial for obtaining local interpretations tailored to a particular input sequence.

\begin{figure}[ht]
    \centering
    \includegraphics[width=0.9\textwidth]{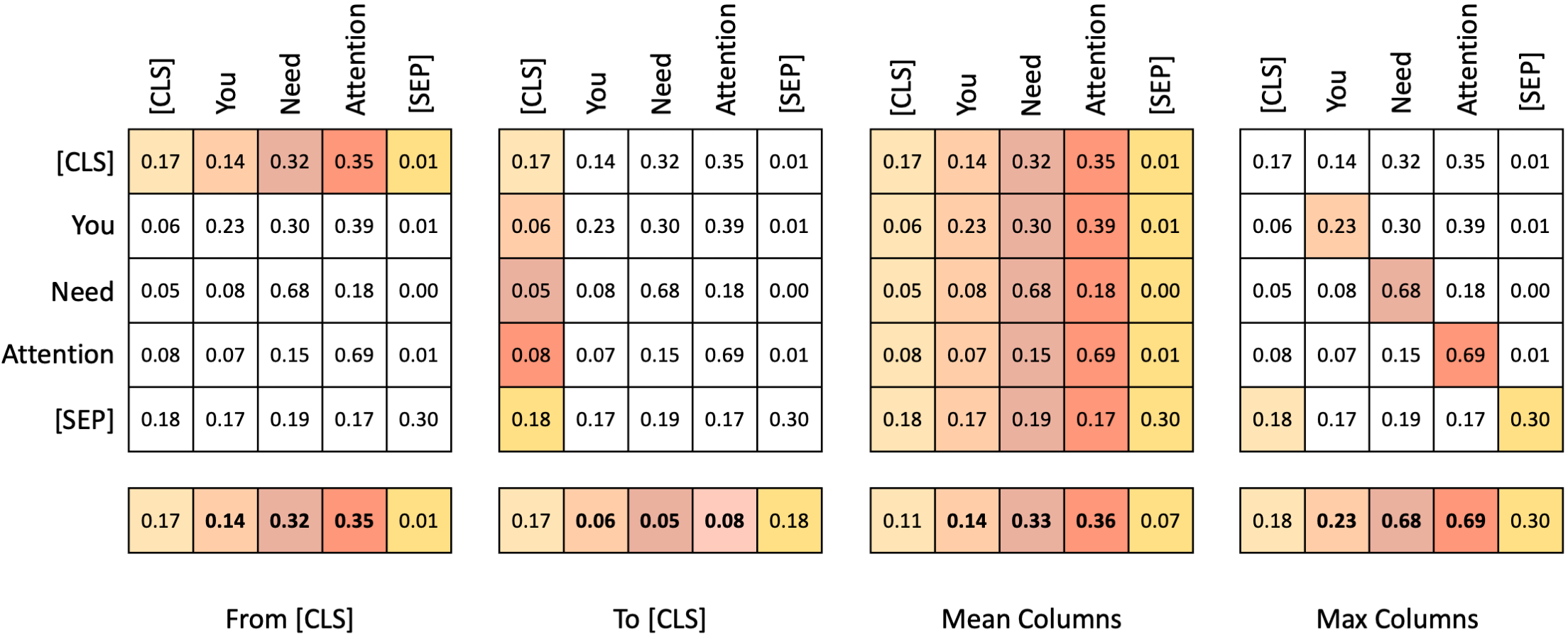}
    \caption{Interpretation extraction operations from an attention matrix}
    \label{fig:attention_vis}
\end{figure}

These head and layer operations lead to a single matrix, integrating the attention scores from the different heads of the different layers in a model. 
To obtain the final interpretation vector, a common approach is to consider the attention that each input token receives from the special [CLS] token that is prepended at the beginning of sequences in text classification tasks~\cite{chefer2021transformer,hateXplain}. We call this operation ``From [CLS]''. Assuming that each row of the attention matrix corresponds to the attention a token pays towards the others, while each column corresponds to the attention it receives from the others, this operation amounts to extracting the [CLS] row of the final attention matrix. We also consider a ``To [CLS]'' operation, by extracting the [CLS] column of the final matrix, containing the attention that the [CLS] token receives from each input token. Two additional operations that were identified in the literature, are selecting the maximum value from each column ``Max Columns''~\cite{schwenke2021show} and averaging the columns of the attention matrix ``Mean Columns''~\cite{clarko}. All these four operations are presented in Figure~\ref{fig:attention_vis}.

\begin{figure}[ht]
    \centering
    \includegraphics[width=0.90\textwidth]{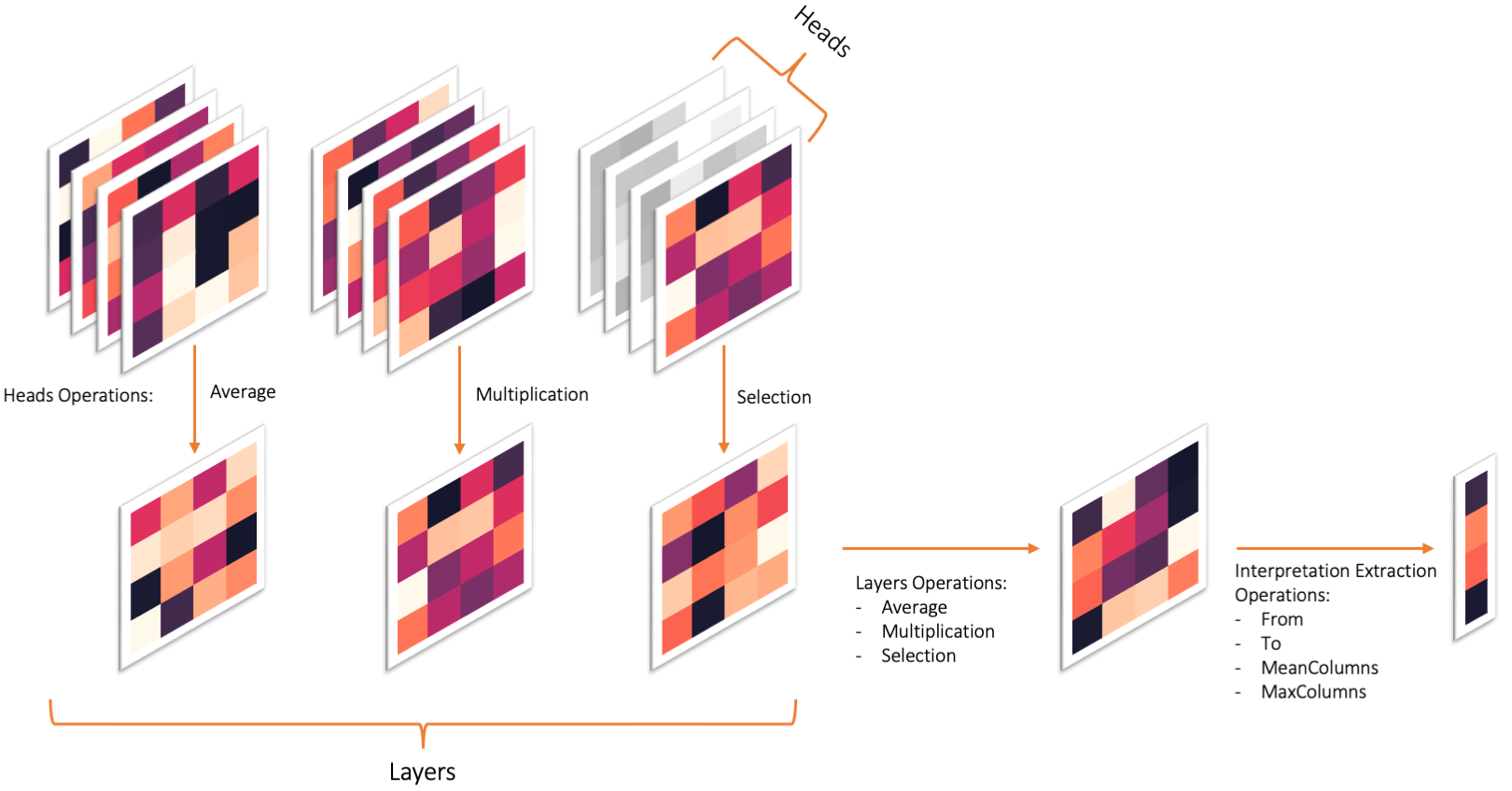}
    \caption{Head, layer and matrix operations}
    \label{fig:attention_op}
\end{figure}

In summary, we considered the following operations that are also depicted in Figure~\ref{fig:attention_op}: a) averaging, multiplying and selection for heads, b) averaging, multiplying and selection for layers, and c) From [CLS], To [CLS], Max Columns and Mean Columns at the matrix level. The combinations of these operations lead to a total of ($2+H$) $\times$ ($2+M$) $\times$ $4$ potentially different attention-based interpretations, where $H$ denotes the number of heads and $M$ the number of layers.  

\begin{algorithm}[ht]
\caption{ExtractInterpretations}
\label{alg:compute_scores}
\begin{algorithmic}[1]
\Require{model $f$, instance $x$}
\State $M$ = number of layers in $f$, $H$ = number of heads in each layer of $f$
\State $A_{ij}$ = attention matrix of layer $i$ and head $j$ in $f$ given $x$
\For{head operation $h \in [mean, multi, 1, \dots, H]$}
    \FFor{$i \in [1, \ldots, M]$}
        $B_i=h(A_i)$ \Comment{apply $h$ across the head matrices of each layer}
    \EndFFor
    \For{layer operation $l \in [mean, multi, 1, \dots, M]$}
        \State $C=l(B)$ 
        \Comment{apply $l$ across the layer matrices}
        \For{matrix operation $m \in [from, to, mean\_col, max\_col]$}
            \State $w[h,l,m]=m(C)$ \Comment{extract interpretation}
        \EndFor
    \EndFor
\EndFor
\State \Return $w$
\end{algorithmic}
\end{algorithm}

\begin{algorithm}[ht]
\caption{\textsc{Optimus Prime}}
\label{alg:maxperinstance}
\begin{algorithmic}[1]
\Require{model $f$, instance $x$, faithfulness evaluation metric $e$}
\State $w = ExtractInterpretations(f,x)$, $best = [mean, mean, from]$, $best\_score = 0$
\For{head operation $h \in [mean, multi, 1, \dots, H]$}
    \For{layer operation $l \in [mean, multi, 1, \dots, M]$}
        \For{matrix operation $m \in [from, to, mean\_col, max\_col]$}
                \State $score = e(w[h,l,m],f,x)$ 
                \IIf{$score > best\_score$}
                    $best\_score = score$, $best = [h,l,m]$
                \EndIIf
        \EndFor
    \EndFor
\EndFor
\State \Return $best$
\end{algorithmic}
\end{algorithm}

\begin{algorithm}[ht]
    \caption{\textsc{Optimus Batch}}
    \label{alg:maxacross}
    \begin{algorithmic}[1]
        \Require model $f$, set of instances $X$, faithfulness evaluation metric $e$
        \State $best = [mean, mean, from]$, $best\_score = 0$        
        \FFor{$i \in [1, \ldots, \vert X \vert]$}
             $w_i = ExtractInterpretations(f,x_i)$
        \EndFFor       
        \For{head operation $h \in [mean, multi, 1, \dots, H]$}
            \For{layer operation $l \in [mean, multi, 1, \dots, M]$}
                \For{matrix operation $m \in [from, to, mean\_col, max\_col]$}
                        \State $score[h,l,m] = 0$
                        \FFor{$i \in [1, \ldots, \vert X \vert]$}
                            $score[h,l,m] = score[h,l,m] + e(w_i[h,l,m],f,x_i)$
                        \EndFFor
                    \IIf{$scores[h,l,m] > best\_score$}
                        $best\_score = scores[h,l,m]$, $best = [h,l,m]$
                    \EndIIf
                \EndFor
            \EndFor
    \EndFor
    \State \Return $best$
    \end{algorithmic}
\end{algorithm}

\subsubsection{Selecting the most faithful interpretation}

Given a transformer model $f$, an input sequence $x$, \textsc{Optimus Prime} extracts interpretations using all combinations of the available operations at the head, layer and matrix level (Algorithm \ref{alg:compute_scores}). To arbitrate among all these interpretations, it uses an unsupervised faithfulness evaluation metric $e$, in order to output the most faithful interpretation (Algorithm \ref{alg:maxperinstance}).  

The first variation, \textsc{Optimus Batch}, finds the most faithful combination of operations for a set of instances (Algorithm~\ref{alg:maxacross}). This combination is then used to provide interpretations for future instances. \textsc{Optimus Batch} is faster and more environmentally friendly because the search for the most faithful combination occurs only once. However, it is expected to yield lower results than \textsc{Optimus Prime}, which performs the process separately for each instance.

Selecting one combination to extract an interpretation in a multi-label task with $L$ labels, where each instance relates to more than one label may be insufficient, as the positive prediction of different labels may have different interpretation. Therefore, \textsc{Optimus Label} acquires multiple interpretations, one for each different label predicted for an examined instance $x$ (see Algorithm~\ref{alg:maxperinstanceperlabel}). While in a binary classification task this makes no difference, in multi-label tasks it can enhance the performance, as there is more flexibility to match the different ground truth rationales of the positive labels.

\begin{algorithm}[ht]
\caption{\textsc{Optimus Label}}
\label{alg:maxperinstanceperlabel}
\begin{algorithmic}[1]
\Require{model $f$, instance $x$, faithfulness evaluation metric $e$}
\State $w = ExtractInterpretations(f,x)$
\For{predicted label $p$ in $f(x)$}
    \State $best_p = [mean, mean, from]$
    \State $best\_score_p = 0$
    \For{head operation $h \in [mean, multi, 1, \dots, H]$}
        \For{layer operation $l \in [mean, multi, 1, \dots, M]$}
            \For{matrix operation $m \in [from, to, mean\_col, max\_col]$}
                    \State $score = e(w[h,l,m],f,x,p)$ 
                    \IIf{$score > best\_score_p$}
                        $best\_score_p = score$, $best_p = [h,l,m]$ 
                    \EndIIf
            \EndFor
        \EndFor
    \EndFor
\EndFor
\State \Return $best$
\end{algorithmic}
\end{algorithm}

\subsection{Ranked Faithful Truthfulness}

The selection process presented in the previous subsection relies heavily on the use of a faithfulness metric. While metrics like faithfulness score~\cite{faithfulness} and truthfulness~\cite{mollas_lionets_2022} could be used, this work introduces a novel feature importance metric named Ranked Faithful Truthfulness (RFT). Inspired by both faithfulness score and truthfulness, this metric combines their qualities to provide a more complete evaluation. RFT examines the whole interpretation, assigning each token a different penalty based on its importance.

Considering model $f$, input sequence $x$ of size $S$, and the interpretation $z$ of $f(x)$, as discussed in the beginning of Section~\ref{sec:opt}, 
RFT performs $S$ independent modifications to $x$, each time removing a different token, $t_i$, leading to instance $x^{(-i)}$. For each modification, it computes a faithfulness score $v$ based on $w_i$ and the difference of $f_p(x)$ and $f_p(x^{(-i)})$, where $f_p$ returns the probability of $x$ belonging in a certain label, as follows: 

\begin{equation}
  \centering    
  \label{eq:compare}
  \textrm{\textit{v}}(x,z,i) = 
  \begin{cases}
    f_p(x) - f_p(x^{(-i)}), & \text{If } w_{i} > 0,\\ 
    f_p(x^{(-i)}) - f_p(x), & \text{If } w_{i} < 0,\\ 
    -\vert f_p(x) - f_p(x^{(-i)}) \vert, & \text{If } w_{i} = 0\\ 
    \end{cases}
\end{equation}

For non-zero weights, this score is positive (negative) when the change in prediction aligns (contrasts) with our expectations given the weight of the model. For zero weights, it is negative or zero. In all cases, its magnitude corresponds to the absolute value of the difference in predictions.

In addition, RFT normalizes this score proportionally to the importance of each token. An intuitive way to achieve this would be to multiply it by the absolute value of the token importance $ \vert w_i \vert$. However, this would result in information loss, as the prediction changes of zero weights would not be considered. We instead divide the score by the rank $r(t_i)$ of token $t_i$ based on the absolute value of its weight. For example, the ranks of 3 tokens with importance values -0.1, 0.3, 0.2, would be 3, 1 and 2, respectively. 
Eq.~\ref{eq:rft} provides the definition of RFT. Higher RFT values indicate better performance.

\begin{equation}
    \label{eq:rft}
    \textrm{RFT}(x,z) = \frac{1}{S}\sum_{i=1}^{S}\frac{v(x,z,i)}{r(t_i)}
\end{equation}

\subsubsection{Token replacement by [UNK]}
\label{sec:unk}

Faithfulness-oriented metrics, including RFT, evaluate the performance of an interpretability technique based on how the model's decision changes when one or more tokens of the input is removed. This however affects the context for the rest of the tokens, which is important in sequence processing models, like recurrent neural networks and transformers, and even more so, if they use positional encoding, which is the standard for transformer models. This is even more apparent when measuring the distributional shift between the original text and the one altered after removing the token~\cite{granularity}. This finding suggests that simply removing words or tokens produces texts that are out-of-distribution for the transformer, greatly hindering the model's performance.

To address this issue, we propose to replace tokens with [UNK] instead of deleting them. This way, we nullify the influence of the replaced token, while minimally affecting the context. Another similar option would be to use [MASK] which the transformer is already familiar with, rather than [UNK]~\cite{liuLGKL022}. However, this would also lead to erroneously deflated scores, since the model is trained to replace [MASK] with words fitting to the context.

\begin{figure}[ht]
\centering
\minipage{0.275\textwidth}
  \includegraphics[width=\linewidth]{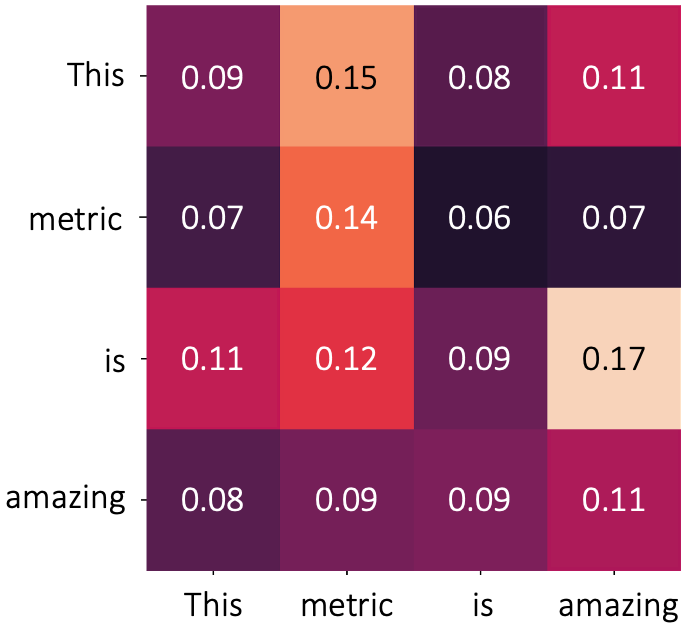}
  \subcaption{Original attention scores}\label{fig:h2}
\endminipage\hfill
\minipage{0.275\textwidth}
  \includegraphics[width=\linewidth]{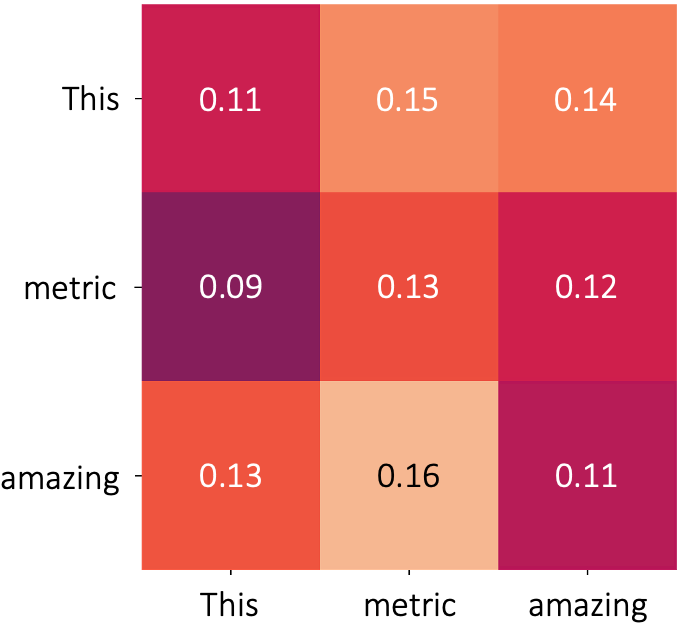}
  \subcaption{After removal of token \textit{is}}\label{fig:h1}
\endminipage\hfill
\minipage{0.33\textwidth}
  \includegraphics[width=\linewidth]{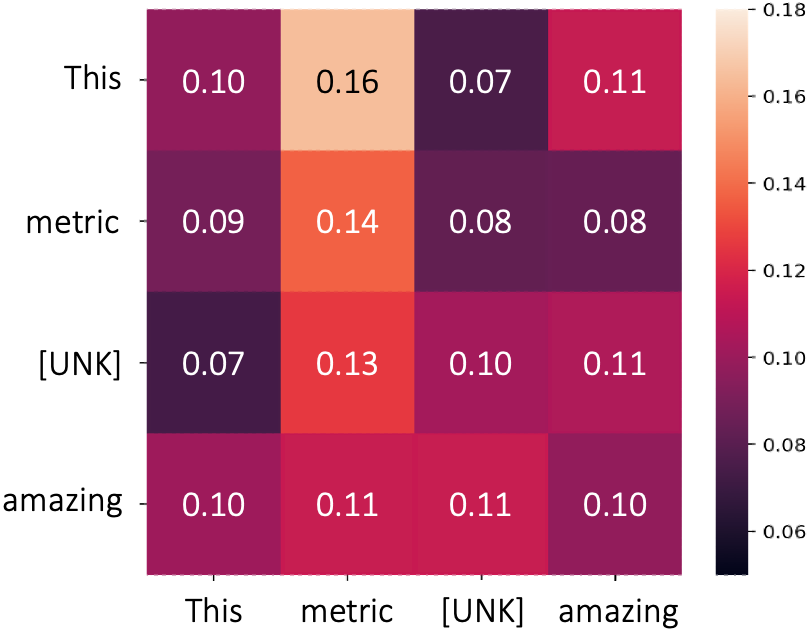}
  \subcaption{After replacement with [UNK]}\label{fig:h3}
\endminipage
\caption{Example of attention with token removal and replacement with [UNK]}
\label{unk_example}
\end{figure}

Figure \ref{unk_example} shows an example of the change in attentions when using [UNK], where image (a) presents the attentions of the initial sequence, (b) after the most important token is removed, and (c) when replaced with [UNK]. We can see that removing the token affects attentions between the remaining tokens more than replacing it with [UNK]. For example, the attention ``metric'' has towards ``amazing'' is $0.07$ in the original sequence. By removing ``is'' attention increases to $0.12$, which is to be expected since the context of the sequence changed, as these tokens are next to each other. On the other hand, replacing ``is'' with [UNK], increases attention slightly to $0.08$, since this change does not affect the positions of the tokens.

\section{Experiments}
\label{sec:exp}
This section presents our experimental setup, a comparative evaluation of RFT against faithfulness, and a comparative evaluation of the proposed attention-based explanation methods against LIME, IG and the standard attention-based technique in terms of the quality of their explanations (both quantitatively and qualitatively) and their computational requirements. The code used in our experiments, as well as a ready to use tool for interpreting transformers using \textsc{Optimus} and RFT are available in GitHub\footnote{\url{https://tinyurl.com/bdh3v2nw}}.

\subsection{Setup}

We use BERT and DistilBERT transformer models in our experiments. BERT was chosen due to its high prominence in the literature, while DistilBERT as a lighter alternative. The base implementation for both models was selected since it is the most common choice for text classification tasks. We compare the performance of our technique to two state-of-the-art competitors, LIME~\cite{LIME}, and IG~\cite{IG} that were covered in Section~\ref{sec:related-work}.

We experiment on 7 datasets coming from 4 different domains: hate speech, biomedicine, sentiment analysis, and natural language understanding (NLU). We selected datasets of sequence length that could fit into the 512 token capacity of the base implementations of BERT and DistilBERT, ideally being accompanied by (token/sentence-level) rationales and covering both single and multi-label classification tasks. Table \ref{tab:datasets} presents key statistics about each dataset, along with the performance of the two transformer models. 


\begin{table}[ht]
\centering
\caption{Key statistics for each dataset. Information about mean size is presented in token/sentence-level. Performance is measured in terms of $F_1$ macro (\%) for BERT/DistilBERT}
\label{tab:datasets}
\begin{tabular}{rccccccc}
\hline
Dataset     & Rationales      & Samples & Mean Size & Labels & Performance & Domain\\
\hline
HX       & token           & 13.749     & 23.9        & 1      & 87.97/87.61 & Hate speech \\
Ethos    & -               & 433       & 20.4        & 8      & 85.71/79.71 & Hate speech \\
AIS      & -               & 3.024      & 44.8/7.1    & 1      & 98.28/98.96 & Biomedicine\\
HoC      & sentence        & 1.852      & 244/10.1  & 10     & 82.44/78.46 & Biomedicine \\
MV       & token           & 421       & 388/18.7     & 1      & 92.61/93.86 & Sent. analysis \\
HB       & token           & 500       & 18.5         & 6      & 61.10/57.39 & Sent. analysis \\
ESNLI    & token           & 10.000     & 24.4        & 1      & 90.13/90.05 & NLU \\
\hline
\end{tabular}
\end{table}



HateXplain (HX)~\cite{hateXplain} is a single-label and Ethos~\cite{ethos} a multi-label dataset from the hate speech domain. They contain hate speech posts collected from Twitter/Gab and YouTube/Reddit, respectively. The former classifies these posts as hateful, normal, offensive or undecided, out of which we disregard the last two classes. In Ethos, each post is associated with eight labels regarding violence, target and type of hate speech. 

From the biomedicine domain, we use Acute Ischemic Stroke (AIS)~\cite{ais}, a single-label dataset, and Hallmarks of Cancer (HoC)~\cite{hoc}, a multi-label one. The first concerns medical notes of brain MRI scans for acute ischemic stroke. HoC, contains biomedical abstracts indexed with 10 hallmarks of cancer. 

From the sentiment analysis domain, two datasets, Movies (MV)~\cite{eraser} and Hummingbird (HB)~\cite{hummingbird}, were employed for single and multi-label tasks, respectively. Movies contains reviews with positive or negative sentiment. Due to the larger sequences of the samples in this dataset, we applied a filtering step, only keeping the ones with size less than or equal to 512. Hummingbird concerns a textual style classification task, with labels including politeness, sentiment, offensiveness, and five emotions. 

ESNLI~\cite{esnli} is a single-label classification dataset for natural language understanding. Given two sentences, the premise and hypothesis, the objective is to determine their relationship: entailment, contradiction, or neutral. Due to the ambiguous nature of the \textit{neutral} class in the examples, we only kept the ones related to the \textit{entailment} and \textit{contradiction} classes. Furthermore, we limit the number of examples to the first 10,000. 


We used 20\% of each dataset as test set, while the remaining 70\% and 10\% were used as training and validation sets, respectively, for fine-tuning the models. We ran our experiments once for each dataset, as all examined techniques besides LIME are deterministic, and the computational complexity of LIME makes multiple runs prohibitive. 



\subsection{Evaluating RFT}

In this experiment, we focus on the five datasets where rationales are available and measure the Pearson and Spearman correlation of the faithfulness score (F) and our RFT metric with the supervised AUPRC metric, which uses ground-truth rationales. For each dataset, we measure the F, RFT and AUPRC performance of each interpretation technique (for \textsc{Optimus} we separately consider each combination of head, layer and matrix operation), by averaging the scores of the techniques across the test instances. Then, we compute the correlation values between the performance of F and AUPRC as well as RFT and AUPRC across the techniques. 

Table~\ref{tab:corr} shows that RFT is more correlated to AUPRC than F. Specifically, in 4 out of the 5 examined datasets RFT has higher correlation to AUPRC than F when evaluating the interpretations for BERT's decisions, while in HoC the 2 metrics have the same correlation. For DistilBERT, we observe a similar pattern, with RFT being more aligned to AUPRC in 4 out of 5 cases and the same in HB, with respect to F. Therefore, we use RFT as the faithfulness evaluation metric of \textsc{Optimus}, as well as for the comparison of the interpretability techniques in the following section.

\begin{table}[ht]
\centering
\caption{Pearson and Spearman correlations between faithfulness score (F) and RFT with AUPRC. Higher correlation denoted with bold}
\label{tab:corr}
\begin{tabular}{rcccccccc}
        & \multicolumn{4}{c}{BERT}                                   & \multicolumn{4}{c}{DistilBERT}                             \\ \cline{2-9} 
        & \multicolumn{2}{c}{Pearson} & \multicolumn{2}{c}{Spearman} & \multicolumn{2}{c}{Pearson} & \multicolumn{2}{c}{Spearman} \\ \cline{2-9} 
        &  F    & RFT  &  F   & RFT  &  F   & RFT  &  F   & RFT  \\ \hline
HX      & 0.93 & \textbf{0.96} & 0.93 & \textbf{0.96} & 0.91 & \textbf{0.96} & 0.93 & \textbf{0.95} \\
HoC & \textbf{0.67} & \textbf{0.67} & \textbf{0.58} & \textbf{0.58} & 0.82 & \textbf{0.85} & 0.82 & \textbf{0.86} \\
MV      & 0.33 & \textbf{0.51} & 0.31 & \textbf{0.40} & 0.67 & \textbf{0.77} & 0.37 & \textbf{0.52} \\
HB      & 0.69 & \textbf{0.76} & 0.67 & \textbf{0.76} & \textbf{0.86} & \textbf{0.86} & \textbf{0.83} & \textbf{0.83} \\
ESNLI   & 0.67 & \textbf{0.77} & 0.63 & \textbf{0.75} & 0.67 & \textbf{0.77} & 0.69 & \textbf{0.74} \\ \hline
Average & 0.66 & \textbf{0.73} & 0.62 & \textbf{0.69} & 0.79 & \textbf{0.84} & 0.72 & \textbf{0.78} \\ \hline
\end{tabular}
\end{table}

\subsection{Quantitative results}
\label{sec:quant}

In the tables below, \textit{B} denotes a baseline attention setup, namely \textit{mean} for heads, \textit{mean} for layers, and \textit{From [CLS]} at the matrix level. These operations are the ones most commonly found in the literature. Regarding our technique, \textsc{OP} represents \textsc{Optimus Prime}, \textsc{OB} the \textsc{Optimus Batch} variant and \textsc{OL} the \textsc{Optimus Label} one. It is worth noting that, \textsc{OP} and \textsc{OL}, yield the same results in binary classification. 

For datasets with larger sequences, namely AIS, HoC, and MV, the evaluation using RFT was performed both at the token and at the sentence level, with the latter designated by (S) next to the dataset name. 
For the sentence level experiments, the weight of a sentence $s_{k}$ is obtained by computing the average weight of its tokens: $\frac{1}{\vert s_{k} \vert} \sum_{t_{i}\in s_{k}}w_{i}$.

The top part of Table \ref{tab:rft-auprc-bert} summarizes the RFT performance of the examined techniques across the different datasets in relation to the explanations provided for BERT. We can see that even the baseline attention setup achieves competitive performance in all datasets when using $A$ as raw attention with a mean rank of 7.7. Furthermore, our proposed unsupervised process further boosts those results, with \textsc{OP} obtaining a mean rank of 2.4. \textsc{OL} increases the performance in multi-label datasets achieving a mean rank of 1.7. These two techniques, however, are computationally demanding since they require a search step for each instance and label. In contrast, \textsc{OB} which also has performance higher than state-of-the-art (mean rank 5.6), is less costly and friendlier to the environment. The results when using $A^*$, show a slight decline for \textsc{OP} (1.9) and \textsc{OL} (2.9). In addition, the baseline setup has the worst performance for $A^*$, consistently being last (10). Investigating the interpretations provided by the $A^*$ baseline setup, we found that many of them consist solely of negative weights, which is the cause for the low RFT performance. The \textsc{OB} results with $A^*$ are much worse than those with $A$.

\begin{table}[ht]
\centering
\caption{Performance of interpretability techniques in terms of RFT (top) and AUPRC (bottom) when explaining BERT on different datasets. Best performance denoted with bold, second best denoted with underline. The average rank of each technique across the datasets is also available}
\label{tab:rft-auprc-bert}
\begin{tabular}{rcccccccccc}
        &      &    & \multicolumn{4}{c}{A} & \multicolumn{4}{c}{A*} \\\cline{4-11}
Dataset & LIME & IG   & B    & \textsc{OB}    & \textsc{OP}   & \textsc{OL}  & B    & \textsc{OB}    & \textsc{OP}   & \textsc{OL}  \\ \hline
HX      & .180 & .487 & .455 & .465 & \underline{.528} & \underline{.528} & .133 & .453 & \textbf{.548} & \textbf{.548} \\
Ethos   & .483 & .515 & .422 & .444 & .543 & \underline{.620} & .181 & .450 & .498 & \textbf{.635} \\
AIS     & .068 & .079 & .063 & .081 & \textbf{.110} & \textbf{.110} & .001 & .077 & \underline{.105} & \underline{.105} \\
AIS (S) & .132 & .164 & .139 & .185 & \textbf{.202} & \textbf{.202} & .008 & .173 & \underline{.201} & \underline{.201} \\
HoC     & .114 & .307 & .239 & .240 & .349 & \textbf{.417} & -.013 & .208 & .326 & \underline{.399} \\
HoC (S) & .141 & .270 & .244 & .273 & .360 & \underline{.404} &-.132 & .203 & .372 & \textbf{.436} \\
MV      & .011 & .062 & .132 & .137 & \textbf{.242} & \textbf{.242} & .001 & .085 & \underline{.201} & \underline{.201} \\
MV (S)  & .063 & .053 & .110 & .144 & \underline{.293} & \underline{.293} & -.124 & .144 & \textbf{.341} & \textbf{.341}\\
HB      & .356 & .151 & .226 & .269 & .408 & \underline{.429} & -.079 & .259 & .410 & \textbf{.437}\\
ESNLI  & .292 & .277 & .246 & .453 & \textbf{.636} & \textbf{.636} & -.074 & .394 & \underline{.627} & \underline{.627}\\  \hline 
Avg. Rank  & 7.9 & 6.7 & 7.7 & 5.6 & 2.4 & 1.7 & 10.0 & 7.0 & 2.9 & 1.9\\ \hline 
\hline
HX      & .296 & \textbf{.391} & .366 & .374 & .369 & .371 & .358 & \underline{.387} & .367 & .369  \\
HoC & .367 & \textbf{.646} & .554 & \underline{.557} & .511 & .547 & .469 & .421 & .493 & .546 \\
MV      & .139 & .169 & .176 & \underline{.183} & \textbf{.188} & \textbf{.188} & \underline{.183} & .147 & .177 & .177  \\
HB      & \textbf{.506} & .366 & .399 & .483 & .479 & .454 & .391 & \underline{.498} & .456 & .443    \\
ESNLI  & .477 & .446 & .411 &\textbf{.510} & \underline{.500} & \underline{.500} & .433 & .452 & .491 & .491   \\  \hline 
Avg. Rank  & 7.4 & 5.6 & 7.2 & 2.4 & 3.6 & 3.6 & 7.8 & 5.8 & 5.6 & 5.4\\
\hline
\end{tabular}
\end{table}

Similarly, the bottom part of Table \ref{tab:rft-auprc-bert} presents the results for the AUPRC metric in datasets with rationales. The baseline attention yet again outperforms LIME in mean rank, while our unsupervised process increases the obtained AUPRC values in most datasets. Here, however, we can see that the best technique is \textsc{OB} using $A$, with a rank of 2.4, instead of \textsc{OP} and \textsc{OL}, which achieve a mean rank of 3.6. The cause of this phenomenon is that our unsupervised procedure finds the best setup per instance according to RFT, which may not necessarily increase the AUPRC scores. Interpretations provided using $A^*$ seem to be less effective than those of $A$ for the AUPRC metric as well, specifically \textsc{OB}'s ranking goes from 2.4 to 5.8, \textsc{OP}'s from 3.6 to 5.6 and \textsc{OL}'s from 3.6 to 5.4. Nevertheless, the results of $A^*$'s RFT evaluation are much lower than AUPRC's. This is due to the two metrics' distinct natures, with RFT considering and evaluating polarity as well as ranking, while AUPRC only ranking. The $A^*$ baseline setup, however, has a better ranking when evaluated by AUPRC (7.8) when compared to RFT's evaluation (10). In this scenario, where we detected numerous interpretations with exclusively negative values, RFT evaluated them harshly due to their polarity, although AUPRC considered the ranking to be relatively correct.

\begin{table}[ht]
\centering
\caption{Performance of interpretability techniques in terms of RFT (top) and AUPRC (bottom) when explaining DistilBERT on different datasets. Best performance denoted with bold, second best denoted with underline. The average rank of each technique across the datasets is also available}
\label{tab:rft-auprc-distilbert}
\begin{tabular}{rcccccccccc}
        &      &      & \multicolumn{4}{c}{A} & \multicolumn{4}{c}{A*}  \\ \cline{4-11} 
Dataset & LIME & IG   & B    & \textsc{OB}    & \textsc{OP}   & \textsc{OL}  & B    & \textsc{OB}    & \textsc{OP}   & \textsc{OL}  \\ \hline
HX      & .166 & .309 & .293 & .349 & \textbf{.477} & \textbf{.477} & .066 & .355 & \underline{.475} & \underline{.475} \\
Ethos   & .499 & .540 & .471 & .497 & .591 & \underline{.651} & .200 & .482 & .596 & \textbf{.658} \\
AIS     & .064 & .082 & .093 & .095 & \textbf{.116} & \textbf{.116} & .001 & .091 & \underline{.113} & \underline{.113} \\
AIS (S) & .111 & .193 & .207 & .216 & \underline{.223} & \underline{.223} &-.079 & .202 & \textbf{.225} & \textbf{.225} \\
HoC     & .079 & .291 & .216 & .216 & .306 & \textbf{.354} & .003 & .204 & .293 & \underline{.334} \\
HoC (S) & .117 & .266 & .245 & .245	& .326 & \underline{.356} &-.098 & .165 & .337 & \textbf{.376} \\
MV      & .023 & .098 & .073 & .080 & \textbf{.185} & \textbf{.185} & -.003 & .079 & \underline{.173} & \underline{.173} \\
MV (S)  & .069 & .140 & .136 & .159 & \underline{.239} & \underline{.239} &-.103 & .158 & \textbf{.253} & \textbf{.253} \\
HB      & .329 & .167 & .281 & .319 & .399 & \textbf{.406} & .194 & .319 & .394 & \underline{.402} \\
ESNLI  & .413 & .353 & .326 & .377 & \textbf{.619} & \textbf{.619} &-.035 & .356 & \underline{.612} & \underline{.612} \\  \hline 
Rank  & 7.9 & 6.8 & 7.4 & 6.0 & 2.4 & 1.6 & 9.9 & 6.9 & 2.8 & 2.0\\ \hline 
\hline
HX      & .322 & .378 & .342 & \textbf{.402} & .381 & \underline{.392} & .375 & \textbf{.402} & .382 & .387 \\
HoC & .387 & \textbf{.671} & \underline{.607} & \underline{.607} & .525 & .596 & .535 & .442 & .494 & .547 \\
MV      & .147 & .240 & .205 & \textbf{.305} & .238 & .238 & .176 & \underline{.275} & .229 & .229 \\
HB      & \textbf{.514} & .372 & .386 & .470 & .488 & .453 & .388 & \underline{.496} & .473 & .447 \\
ESNLI  & .463 & .447 & .409 &.489 & \underline{.497} & \underline{.497} & .487 & \textbf{.501} & .496 & .496    \\ \hline 
Rank  & 7.8 & 6.0 & 7.6 & 3.0 & 4.4 & 3.8 & 7.6 & 3.2 & 5.4 & 5.2\\
\hline
\end{tabular}
\end{table}

The findings for DistilBERT showcased in Table \ref{tab:rft-auprc-distilbert} are similar to the aforementioned ones, suggesting that our technique is stable across different encoder-based Transformer architectures.

\begin{figure}[ht]
    \centering
    \includegraphics[width=1\textwidth]{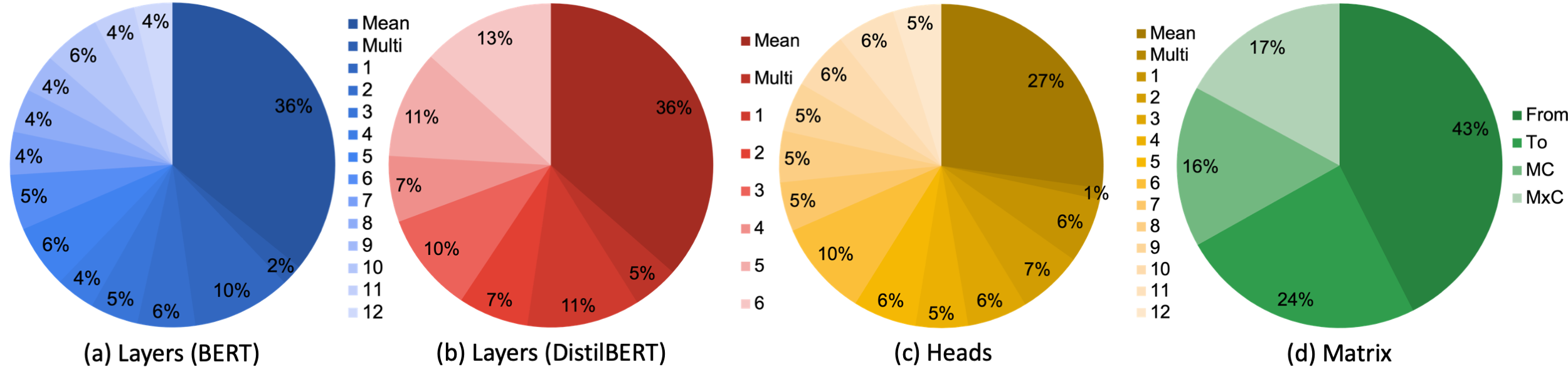}
    \caption{Frequency of operations in layer, head and matrix level}
    \label{fig:operation_pies}
\end{figure}

Another noteworthy discovery is an analysis of the most commonly used options for layer, head, and matrix operations as indicated by \textsc{OP}'s per instance best interpretations based on AUPRC, with both $A$ and $A^*$ (Figure~\ref{fig:operation_pies}). This analysis indicates that, while the most commonly used operations in the literature appear more frequently (mean, multi, From [CLS]), they are insufficient for providing the best attention-based interpretations. As a result, using \textsc{Optimus}, our suggestions (layer and head selection, To [CLS]) can improve those interpretations.

Specifically, Figures~\ref{fig:operation_pies}a and \ref{fig:operation_pies}b show the percentage of occurrence for each operation at the layer level for BERT (12 layers) and DistilBERT (6 layers). We can see that the mean operation is the most common, occurring in 36\% of the interpretations, with multi being selected in only 2\% and 5\% of the interpretations. In case of BERT, the first layer is more commonly employed (10\%), whereas DistilBERT favors the last layer (13\%). In both scenarios, the rest of the layers occur with similar frequency. In Figure~\ref{fig:operation_pies}c, we can see the frequency of operations for heads concerning both BERT and DistilBERT. Similarly to before, mean is the most prevalent (27\%) and multi is the least dominant (1\%). This time, though, all the heads appear at a similar frequency (5-7\%), with only the sixth one standing out (10\%). Finally, among matrix operations (Figure~\ref{fig:operation_pies}d), the most common is From [CLS], with 43\%, followed by To [CLS], with 24\%. Mean Columns (MC) and Max Columns (MxC) occur with comparable frequency in BERT and DistilBERT.

\subsection{Qualitative results}

In addition to the quantitative results, we present two examples, one from the domain of hate speech and one from biomedicine, to qualitatively evaluate the interpretations provided by LIME, IG, Baseline and \textsc{OL}. Red (blue) highlighting refers to a positive (negative) influence on the prediction, with more intense colors suggesting a larger influence. 

\begin{figure}[ht]
    \centering
    \includegraphics[width=0.9\textwidth]{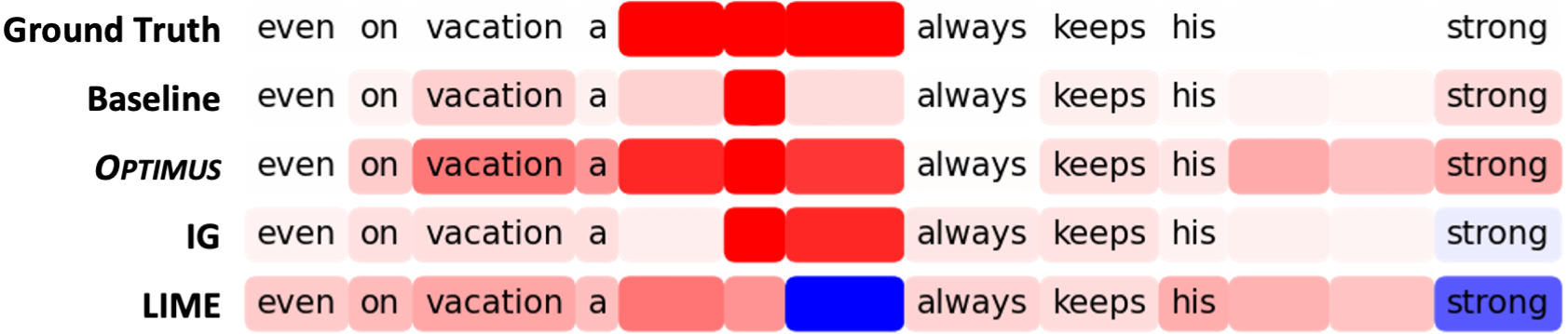}
    \caption{Interpretation example for each examined technique on HX}
    \label{fig:techniques-hate}
\end{figure}

Starting with the hate speech domain, a random instance from the HX dataset predicted by DistilBERT as ``\textit{hate speech}'' is selected. In Figure~\ref{fig:techniques-hate}, we showcase the selected instance, having removed any tokens that may correspond to offensive or derogatory words or imply them. The first line corresponds to ground truth rationales for the examined instance. Baseline attention seems to highlight the important tokens in the sequence, however, giving higher importance to only one of them, according to the ground truth. On the other hand, \textsc{Optimus} correctly identifies the most important tokens, giving minimal weight to others. IG behaves similarly, assigning correct values to important tokens and also including a few irrelevant ones. LIME seems to miss some important tokens, even assigning a negative score to one. 

One intriguing aspect of this example is that the two hidden tokens preceding ``\textit{strong}'' are not included in the ground truth rationale, even though they should have been. This is also supported by the interpretations, which assigned an influence score to those tokens. Indeed, removing these two tokens, has a negative impact on the prediction's probability. This suggests that even ground truth information might be prone to inaccuracies or annotator bias.

\begin{figure}[ht]
    \centering
    \includegraphics[width=\textwidth]{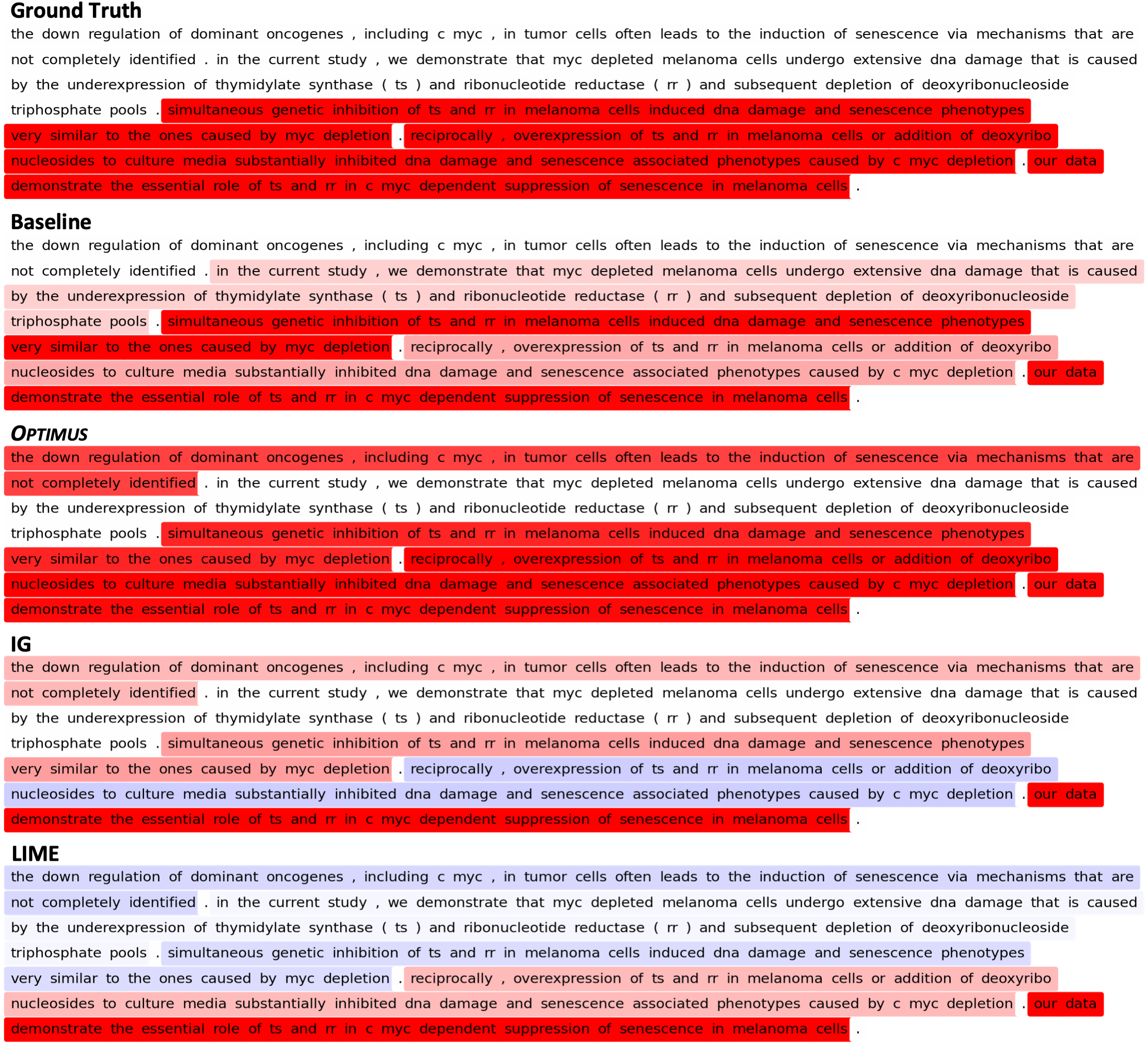}
    \caption{Interpretation example for each examined technique on HoC}
    \label{fig:techniques-bio}
\end{figure}

Having presented a token-level example, the second one concerns sentence-level interpretations. As such, we select HoC, where sentence-level rationales are available. Choosing a random instance, we obtain its prediction, and select one label among the predicted, namely \textit{enabling replicative immortality}. Figure~\ref{fig:techniques-bio} illustrates the interpretations provided by the techniques for that instance and label. Baseline correctly highlights all three important sentences, but gives less importance to one of them, while also highlighting an unrelated one, according to the ground truth. \textsc{OL} accurately identifies all three sentences. However, it assigns importance weight to an unrelated one as well. IG and LIME both erroneously give one of the three ground truth sentences a negative importance score, but correctly underline the other two.

\subsection{Computational overhead analysis} 


One key advantage of using attention interpretations, is the low computational overhead since they are already computed during inference. This leads to faster response times and lower environmental impact. This is especially true in cases where one combination of operations is identified beforehand (B and \textsc{OB} as seen in Section~\ref{sec:quant}) and then applied on incoming instances. In contrast, identifying the most faithful combination for each instance individually (\textsc{OP} and \textsc{OL} as seen in Section~\ref{sec:quant}) and techniques such as LIME and IG require additional procedures, resulting in increased computational cost. 

To empirically validate this, we performed a time response analysis for each technique in HX (small sequences), HoC and MV (large sequences) using BERT. Additionally, we computed their carbon footprint in terms of metric tons of CO$_2$ equivalent emissions (tCO$_2$e) based on~\cite{carbonem}. We performed our experiments in Google Colab using a single GPU with an average power consumption of 271W. We used the formula tCO$_2$e$=$KWh$\times$kg CO$_2$e per KWh$/10^3$, assuming kg CO$_2$e/KWh to be 0.429~\cite{carbonem} and the formula KWh$=($interpretation time$\times \#$ of GPUs $\times$ avg. power/GPU $\times$ PUE)$/10^3$, assuming PUE\footnote{\url{https://tinyurl.com/2u8zeks8}} (power usage effectiveness) to be 1.10. The average tCO$_2$e per instance for each technique can be found in Table~\ref{tab:tco2e}.

\begin{table}[ht]
\centering
\caption{Average time response and tCO$_2e$ emissions}
\label{tab:tco2e}
\begin{tabular}{rcccccc}
         & \multicolumn{2}{c}{HX} & \multicolumn{2}{c}{HoC (S)}  & \multicolumn{2}{c}{MV} \\ \hline
         & Seconds  & tCO2e    & Seconds & tCO2e  & Seconds    & tCO2e    \\ \hline
LIME     & 38.252   & 4.892    & 134.903 & 17.252 & 27.422     & 3.507    \\
IG       & 0.393    & 0.050    & 14.699  & 1.880  & 3.451      & 0.441    \\
B/\textsc{OB}      & 4.21E-05 & 5.38E-06 & 0.010   & 0.001  & 0.009      & 0.001    \\
\textsc{OP}/\textsc{OL}   & 3.050    & 0.390    & 5.304   & 0.678  & 132.204    & 16.907   \\ \hline
\end{tabular}
\end{table}

The average time response (seconds) and tCO$_2e$ emissions for each examined technique are showcased in Table~\ref{tab:tco2e}. When interpretations are provided at sentence-level, \textsc{OP} and \textsc{OL} seem to have lower time responses and emissions, compared to LIME and IG, while at token-level the opposite holds true. This is due to RFT, which is an important part of the \textsc{OP} and \textsc{OL} procedures, needing to examine fewer interpretation elements at the sentence-level. It is worth mentioning that in the case of LIME, we used 2000 neighbors in HX, and 200 in HoC (S) and MV, as LIME cannot be efficiently applied in datasets with larger sequences. Finally, B and \textsc{OB} require the least amount of time. 

\begin{figure}[ht]
    \centering
    \includegraphics[width=1\textwidth]{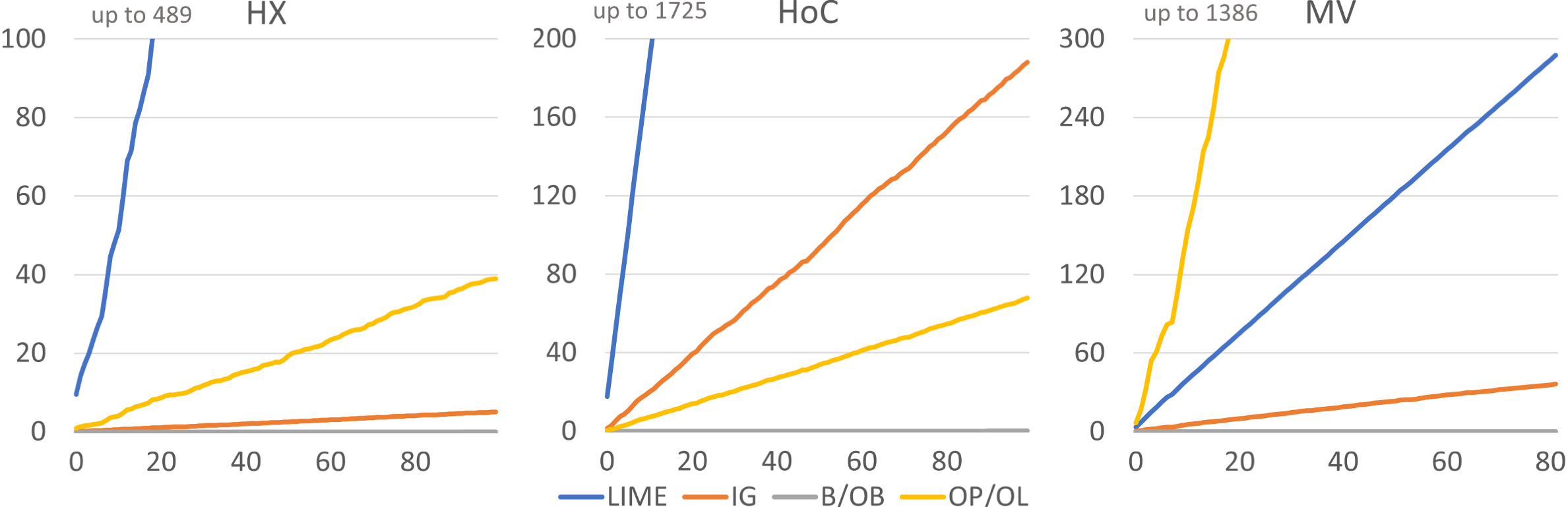}
    \caption{Interpolated emissions for each method (Left: HX, Middle: HoC (S), Right: MV). $X$-axis: number of instances, $Y$-axis: cumulative tCO$_2$e emissions}
    \label{fig:cost}
\end{figure}

Figure~\ref{fig:cost} presents the cumulative tCO$_2$e emissions for up to 100 instances. On the left, we simulate an interpretable Hate Speech detection system on a social media platform, where a large amount of data is being produced every second. Likewise, the middle plot concerns interpretability-assisted semantic indexing of biomedical publications, which tend to have bigger sequences, in conjunction to the large number of publications processed each day in databases like PubMed (986K articles for 2020\footnote{\url{https://tinyurl.com/4nprsskn}}). Finally, on the right, we model a database in which users constantly search for reviews, and an interpretable sentiment analysis system provides insight to these reviews.

\section{Conclusions and Future Work}
\label{sec:con}

Some studies argue that attention should not be used as an interpretation tool, while others incorporate attention-based methods in their experiments without specifying how interpretations are produced. This work investigates various ways attention is used in the literature as interpretation and proposes an arbitration scheme to determine the best way to extract interpretations from attention information. The most faithful combination is determined via an optimization procedure based on an unsupervised faithfulness metric. 

Our findings support that, when properly configured, especially using operations that select particular attention matrices from heads and layers, attention can effectively be used as an interpretation tool for text classification. In addition, we demonstrate that attention can compete with other cutting-edge techniques in a series of experiments that include a new faithfulness metric based on feature importance. Furthermore, compared to other techniques, attention is easier to implement, while also being faster and less harmful to the environment, in certain cases, such as B and \textsc{Optimus Batch}. Despite the effectiveness of attention, identifying the most faithful combination of operations per instance and per label (\textsc{Optimus Prime} and \textsc{Optimus Label}), which achieves the highest performance, according to our experiments is not time efficient.

In the future, we will explore how to reduce the runtime of the \textsc{Optimus Prime} and \textsc{Optimus Label} setups to make them more environmentally friendly. One such way would be to experiment with other faithfulness metrics, that are more time efficient or different unsupervised metrics (e.g., robustness, complexity). Another approach would be to explore the differences in performance between using both $A$ and $A^*$ as input to \textsc{Optimus} techniques instead of using only one of them to extract interpretations. Finally, since both examined transformer models, BERT and DistilBERT, are encoder based, experiments investigating encoder-decoder based models could also be conducted.

\section*{Acknowledgements}
The research work was supported by the Hellenic Foundation for Research and Innovation (H.F.R.I.) under the “First Call for H.F.R.I. Research Projects to support Faculty members and Researchers and the procurement of high-cost research equipment grant” (Project Number: 514)

\bibliographystyle{unsrt}  

\begin{thebibliography}{10}

\bibitem{vaswani2017attention}
Ashish Vaswani, Noam Shazeer, Niki Parmar, Jakob Uszkoreit, Llion Jones,
  Aidan~N Gomez, {\L}ukasz Kaiser, and Illia Polosukhin.
\newblock Attention is all you need.
\newblock {\em Advances in neural information processing systems}, 30, 2017.

\bibitem{wolf-etal-2020-transformers}
Thomas Wolf, Lysandre Debut, Victor Sanh, Julien Chaumond, Clement Delangue,
  Anthony Moi, Pierric Cistac, Tim Rault, Remi Louf, Morgan Funtowicz, Joe
  Davison, Sam Shleifer, Patrick von Platen, Clara Ma, Yacine Jernite, Julien
  Plu, Canwen Xu, Teven Le~Scao, Sylvain Gugger, Mariama Drame, Quentin Lhoest,
  and Alexander Rush.
\newblock Transformers: State-of-the-art natural language processing.
\newblock In {\em Proceedings of EMNLP 2020: System Demonstrations}, pages
  38--45, Online, October 2020. ACL.

\bibitem{transformer_not_inter}
Leonid Schwenke and Martin Atzmueller.
\newblock Show me what you’re looking for: Visualizing abstracted transformer
  attention for enhancing their local interpretability on time series data.
\newblock {\em The International FLAIRS Conference Proceedings}, 34, Apr. 2021.

\bibitem{highrisk2}
EU.
\newblock Proposal for a regulation of the european parliament and the council
  laying down harmonised rules on artificial intelligence (ai act) and amending
  certain union legislative acts.
\newblock {\em EUR-Lex-52021PC0206}, 2021.

\bibitem{notAtt}
Sarthak Jain and Byron~C. Wallace.
\newblock Attention is not explanation.
\newblock In {\em {NAACL-HLT}}, pages 3543--3556, Minneapolis, Minnesota, 2019.
  ACL.

\bibitem{chefer2021transformer}
Hila Chefer, Shir Gur, and Lior Wolf.
\newblock Transformer interpretability beyond attention visualization.
\newblock In {\em Proceedings of the IEEE/CVF Conference on Computer Vision and
  Pattern Recognition (CVPR)}, pages 782--791, June 2021.

\bibitem{DBLP:journals/corr/abs-2005-00928}
Samira Abnar and Willem~H. Zuidema.
\newblock Quantifying attention flow in transformers.
\newblock {\em CoRR}, abs/2005.00928, 2020.

\bibitem{LIME}
Marco~Tulio Ribeiro, Sameer Singh, and Carlos Guestrin.
\newblock Why should i trust you?: Explaining the predictions of any
  classifier.
\newblock In {\em Proceedings of the 22nd ACM SIGKDD international conference
  on knowledge discovery and data mining}, pages 1135--1144. ACM, 2016.

\bibitem{SHAP}
Scott~M Lundberg and Su-In Lee.
\newblock A unified approach to interpreting model predictions.
\newblock In {\em Advances in Neural Information Processing Systems 30}, pages
  4765--4774. Curran Associates, Inc., Long Beach, California, 2017.

\bibitem{LRP}
Sebastian Bach, Alexander Binder, Grégoire Montavon, Frederick Klauschen,
  Klaus-Robert Müller, and Wojciech Samek.
\newblock On pixel-wise explanations for non-linear classifier decisions by
  layer-wise relevance propagation.
\newblock {\em PLOS ONE}, 10(7):1--46, 07 2015.

\bibitem{IG}
Mukund Sundararajan, Ankur Taly, and Qiqi Yan.
\newblock Axiomatic attribution for deep networks.
\newblock In {\em Proceedings of the 34th International Conference on Machine
  Learning, {ICML} 6-11 August}, volume~70, pages 3319--3328, Sydney, NSW,
  Australia, 2017. {PMLR}.

\bibitem{hateXplain}
Binny Mathew, Punyajoy Saha, Seid~Muhie Yimam, Chris Biemann, Pawan Goyal, and
  Animesh Mukherjee.
\newblock Hatexplain: {A} benchmark dataset for explainable hate speech
  detection.
\newblock In {\em Thirty-Fifth {AAAI} Conference on Artificial Intelligence,
  February 2-9}, pages 14867--14875, Online, 2021. {AAAI} Press.

\bibitem{Feldhus2021ThermostatAL}
Nils Feldhus, Robert Schwarzenberg, and Sebastian Moller.
\newblock Thermostat: A large collection of nlp model explanations and analysis
  tools.
\newblock In {\em EMNLP}, 2021.

\bibitem{alammar-2021-ecco}
J~Alammar.
\newblock Ecco: An open source library for the explainability of transformer
  language models.
\newblock In {\em Proceedings of the 59th Annual Meeting of the ACL and the
  11th International Joint Conference on Natural Language Processing: System
  Demonstrations}, pages 249--257, Online, August 2021. ACL.

\bibitem{attExp1}
James Mullenbach, Sarah Wiegreffe, Jon Duke, Jimeng Sun, and Jacob Eisenstein.
\newblock Explainable prediction of medical codes from clinical text.
\newblock In {\em {NAACL-HLT}}, pages 1101--1111, New Orleans, Louisiana, 2018.
  ACL.

\bibitem{attExp2}
Sarah Wiegreffe and Yuval Pinter.
\newblock Attention is not not explanation.
\newblock In {\em {EMNLP/IJCNLP}}, pages 11--20, Hong Kong, China, 2019. ACL.

\bibitem{notAtt2}
Jasmijn Bastings and Katja Filippova.
\newblock The elephant in the interpretability room: Why use attention as
  explanation when we have saliency methods?
\newblock In {\em BlackboxNLP@EMNLP}, pages 149--155, Online, 2020. ACL.

\bibitem{ijcai2022p102}
Runliang Niu, Zhepei Wei, Yan Wang, and Qi~Wang.
\newblock Attexplainer: Explain transformer via attention by reinforcement
  learning.
\newblock In {\em Proceedings of the 31st International Joint Conference on
  Artificial Intelligence, {IJCAI-22}}, pages 724--731, Vienna, Austria, 7
  2022.

\bibitem{electronics10182195}
Luca Bacco, Andrea Cimino, Felice Dell’Orletta, and Mario Merone.
\newblock Explainable sentiment analysis: A hierarchical transformer-based
  extractive summarization approach.
\newblock {\em Electronics}, 10(18), 2021.

\bibitem{DBLP:journals/corr/abs-1906-05714}
Jesse Vig.
\newblock A multiscale visualization of attention in the transformer model.
\newblock {\em CoRR}, abs/1906.05714, 2019.

\bibitem{DBLP:journals/corr/abs-1908-08593}
Olga Kovaleva, Alexey Romanov, Anna Rogers, and Anna Rumshisky.
\newblock Revealing the dark secrets of {BERT}.
\newblock {\em CoRR}, abs/1908.08593, 2019.

\bibitem{DBLP:conf/iclr/BrunnerLPRCW20}
Gino Brunner, Yang Liu, Damian Pascual, Oliver Richter, Massimiliano Ciaramita,
  and Roger Wattenhofer.
\newblock On identifiability in transformers.
\newblock In {\em 8th International Conference on Learning Representations,
  {ICLR}}, Online, 2020. OpenReview.net.

\bibitem{evrikaKatharistiko}
Piyawat Lertvittayakumjorn and Francesca Toni.
\newblock Human-grounded evaluations of explanation methods for text
  classification.
\newblock In {\em Proceedings of the Conference on Empirical Methods in Natural
  Language Processing and the 9th International Joint Conference on Natural
  Language Processing, {EMNLP-IJCNLP}, November 3-7}, pages 5194--5204, Hong
  Kong, China, 2019. ACL.

\bibitem{persilMalaktiko}
Bernease Herman.
\newblock The promise and peril of human evaluation for model interpretability.
\newblock {\em ArXiv}, abs/1711.07414, 2017.

\bibitem{eraser}
Jay DeYoung, Sarthak Jain, Nazneen~Fatema Rajani, Eric Lehman, Caiming Xiong,
  Richard Socher, and Byron~C. Wallace.
\newblock {ERASER}: {A} benchmark to evaluate rationalized {NLP} models.
\newblock In {\em Proceedings of the 58th Annual Meeting of the ACL}, pages
  4443--4458, Online, July 2020. ACL.

\bibitem{robustness}
David~Alvarez Melis and Tommi Jaakkola.
\newblock Towards robust interpretability with self-explaining neural networks.
\newblock In {\em Advances in Neural Information Processing Systems}, pages
  7775--7784, Montreal, Canada, 2018.

\bibitem{comprehensibility_nzw}
Marko Robnik{-}Sikonja and Marko Bohanec.
\newblock Perturbation-based explanations of prediction models.
\newblock In {\em Human and Machine Learning - Visible, Explainable,
  Trustworthy and Transparent}, pages 159--175. Springer, International, 2018.

\bibitem{faithfulness}
Mengnan Du, Ninghao Liu, Fan Yang, Shuiwang Ji, and Xia Hu.
\newblock On attribution of recurrent neural network predictions via additive
  decomposition.
\newblock In {\em The World Wide Web Conference}, pages 383--393, 2019.

\bibitem{mollas_lionets_2022}
Ioannis Mollas, Nick Bassiliades, and Grigorios Tsoumakas.
\newblock {LioNets}: a neural-specific local interpretation technique
  exploiting penultimate layer information.
\newblock {\em Applied Intelligence}, May 2022.

\bibitem{chan-etal-2022-comparative}
Chun~Sik Chan, Huanqi Kong, and Liang Guanqing.
\newblock A comparative study of faithfulness metrics for model
  interpretability methods.
\newblock In {\em Proceedings of the 60th Annual Meeting of the ACL (Volume 1:
  Long Papers)}, pages 5029--5038, Dublin, Ireland, May 2022. ACL.

\bibitem{liuLGKL022}
Yibing Liu, Haoliang Li, Yangyang Guo, Chenqi Kong, Jing Li, and Shiqi Wang.
\newblock Rethinking attention-model explainability through faithfulness
  violation test.
\newblock In {\em International Conference on Machine Learning, {ICML}, 17-23
  July}, volume 162, pages 13807--13824, Baltimore, Maryland, 2022. {PMLR}.

\bibitem{9671639}
Shengzhong Liu, Franck Le, Supriyo Chakraborty, and Tarek Abdelzaher.
\newblock On exploring attention-based explanation for transformer models in
  text classification.
\newblock In {\em IEEE International Conference on Big Data (Big Data)}, pages
  1193--1203, 2021.

\bibitem{wanktree}
Yaushian Wang, Hung-Yi Lee, and Yun-Nung Chen.
\newblock Tree transformer: Integrating tree structures into self-attention.
\newblock In {\em Proceedings of EMNLP 2019 and the 9th International Joint
  Conference on Natural Language Processing (EMNLP-IJCNLP)}, pages 1061--1070,
  Hong Kong, China, November 2019. ACL.

\bibitem{exbert}
Benjamin Hoover, Hendrik Strobelt, and Sebastian Gehrmann.
\newblock ex{BERT}: {A} {V}isual {A}nalysis {T}ool to {E}xplore {L}earned
  {R}epresentations in {T}ransformer {M}odels.
\newblock In {\em Proceedings of the 58th Annual Meeting of the ACL: System
  Demonstrations}, pages 187--196, Online, July 2020. ACL.

\bibitem{schwenke2021show}
Leonid Schwenke and Martin Atzmueller.
\newblock Show me what you’re looking for: visualizing abstracted transformer
  attention for enhancing their local interpretability on time series data.
\newblock In {\em The International FLAIRS Conference Proceedings}, volume~34,
  2021.

\bibitem{clarko}
Kevin Clark, Urvashi Khandelwal, Omer Levy, and Christopher~D. Manning.
\newblock What does {BERT} look at? an analysis of {BERT}{'}s attention.
\newblock In {\em BlackboxNLP@EMNLP}, pages 276--286, Florence, Italy, August
  2019. ACL.

\bibitem{granularity}
Yves Rychener, Xavier Renard, Djamé Seddah, Pascal Frossard, and Marcin
  Detyniecki.
\newblock On the granularity of explanations in model agnostic nlp
  interpretability, 2020.
\newblock To appear in ECMLPKDD2022 proceedings of XKDD workshop.

\bibitem{ethos}
Ioannis Mollas, Zoe Chrysopoulou, Stamatis Karlos, and Grigorios Tsoumakas.
\newblock {ETHOS}: a multi-label hate speech detection dataset.
\newblock {\em Complex \& Intelligent Systems}, January 2022.

\bibitem{ais}
Chulho Kim, Vivienne Zhu, Jihad Obeid, and Leslie Lenert.
\newblock Natural language processing and machine learning algorithm to
  identify brain mri reports with acute ischemic stroke.
\newblock {\em PLOS ONE}, 14(2):1--13, 02 2019.

\bibitem{hoc}
Simon Baker, Ilona Silins, Yufan Guo, Imran Ali, Johan Högberg, Ulla Stenius,
  and Anna Korhonen.
\newblock {Automatic semantic classification of scientific literature according
  to the hallmarks of cancer}.
\newblock {\em Bioinformatics}, 32(3):432--440, 10 2015.

\bibitem{hummingbird}
Shirley~Anugrah Hayati, Dongyeop Kang, and Lyle Ungar.
\newblock Does {BERT} learn as humans perceive? understanding linguistic styles
  through lexica.
\newblock In {\em Proceedings of the 2021 Conference on Empirical Methods in
  Natural Language Processing, {EMNLP}, 7-11 November}, pages 6323--6331,
  Online, 2021. ACL.

\bibitem{esnli}
Oana-Maria Camburu, Tim Rockt{\"a}schel, Thomas Lukasiewicz, and Phil Blunsom.
\newblock e-snli: Natural language inference with natural language
  explanations.
\newblock {\em Advances in Neural Information Processing Systems}, 31, 2018.

\bibitem{carbonem}
David Patterson, Joseph Gonzalez, Quoc Le, Chen Liang, Lluis-Miquel Munguia,
  Daniel Rothchild, David So, Maud Texier, and Jeff Dean.
\newblock Carbon emissions and large neural network training, 2021.

\end{thebibliography}

\end{document}